# Reading Dependencies from Polytree-Like Bayesian Networks

Jose M. Peña
Division of Computational Biology
Department of Physics, Chemistry and Biology
Linköping University, SE-58183 Linköping, Sweden

## Abstract

We present a graphical criterion for reading dependencies from the minimal directed independence map $G$ of a graphoid $p$ when $G$ is a polytree and $p$ satisfies composition and weak transitivity. We prove that the criterion is sound and complete. We argue that assuming composition and weak transitivity is not too restrictive.

## 1 INTRODUCTION

A minimal directed independence map $G$ of an independence model $p$ is used to read independencies that hold in $p$. Sometimes, however, $G$ can also be used to read dependencies holding in $p$. For instance, if $p$ is a graphoid that is faithful to $G$ then, by definition, lack of vertex separation is a sound and complete graphical criterion for reading dependencies from $G$. If $p$ is simply a graphoid, then there also exists a sound and complete graphical criterion for reading dependencies from $G$ [Bouckaert, 1995]. However, this criterion cannot be applied to check whether two set of nodes $\mathbf{X}$ and $\mathbf{Y}$ are dependant given a third set $\mathbf{Z}$ unless $\mathbf{X}$ and $\mathbf{Y}$ are adjacent in $G$, i.e. unless there is an edge in $G$ from some $A \in \mathbf{X}$ to some $B \in \mathbf{Y}$. In this paper, we try to overcome this shortcoming so that $\mathbf{X}$ and $\mathbf{Y}$ are not required to be adjacent in $G$. We do so by constraining $G$ and $p$. Specifically, we present a sound and complete graphical criterion for reading dependencies from $G$ under the assumptions that $G$ is a polytree and $p$ is a graphoid that satisfies composition and weak transitivity. We argue that assuming composition and weak transitivity is not too restrictive. Specifically, we show that there exist important families of probability distributions, among them Gaussian probability distributions, that satisfy these two properties.

The rest of the paper is organized as follows. We start by reviewing some concepts in Section 2. We show in Section 3 that assuming composition and weak transitivity is not too restrictive. We present our graphical criterion in Section 4 and prove that it is sound and complete. Finally, we close with some discussion in Section 5.

## 2 PRELIMINARIES

The definitions and results in this section are taken from [Lauritzen, 1996, Pearl, 1988, Studený, 2005]. We use the juxtaposition $\mathbf{XY}$ to denote $\mathbf{X} \cup \mathbf{Y}$, and $X$ to denote the singleton $\{X\}$. Let $\mathbf{U}$ denote a set of random variables. Unless otherwise stated, all the independence models and graphs in this paper are defined over $\mathbf{U}$.

Let $\mathbf{X}$, $\mathbf{Y}$, $\mathbf{Z}$ and $\mathbf{W}$ denote four mutually disjoint subsets of $\mathbf{U}$. An independence model $p$ is a set of independencies of the form $\mathbf{X}$ is independent of $\mathbf{Y}$ given $\mathbf{Z}$. We represent that an independence is in $p$ by $\mathbf{X} \perp\!\!\!\perp \mathbf{Y}|\mathbf{Z}$ and that an independence is not in $p$ by $\mathbf{X} \not\!\perp\!\!\!\perp \mathbf{Y}|\mathbf{Z}$. In the latter case, we may equivalently say that the dependence $\mathbf{X} \not\!\perp\!\!\!\perp \mathbf{Y}|\mathbf{Z}$ is in $p$. An independence model is a graphoid when it satisfies the following five properties: Symmetry $\mathbf{X} \perp\!\!\!\perp \mathbf{Y}|\mathbf{Z} \Rightarrow \mathbf{Y} \perp\!\!\!\perp \mathbf{X}|\mathbf{Z}$, decomposition $\mathbf{X} \perp\!\!\!\perp \mathbf{YW}|\mathbf{Z} \Rightarrow \mathbf{X} \perp\!\!\!\perp \mathbf{Y}|\mathbf{Z}$, weak union $\mathbf{X} \perp\!\!\!\perp \mathbf{YW}|\mathbf{Z} \Rightarrow \mathbf{X} \perp\!\!\!\perp \mathbf{Y}|\mathbf{ZW}$, contraction $\mathbf{X} \perp\!\!\!\perp \mathbf{Y}|\mathbf{ZW} \wedge \mathbf{X} \perp\!\!\!\perp \mathbf{W}|\mathbf{Z} \Rightarrow \mathbf{X} \perp\!\!\!\perp \mathbf{YW}|\mathbf{Z}$, and intersection $\mathbf{X} \perp\!\!\!\perp \mathbf{Y}|\mathbf{ZW} \wedge \mathbf{X} \perp\!\!\!\perp \mathbf{W}|\mathbf{ZY} \Rightarrow \mathbf{X} \perp\!\!\!\perp \mathbf{YW}|\mathbf{Z}$. Any strictly positive probability distribution is a graphoid.

Let $sep(\mathbf{X}, \mathbf{Y}|\mathbf{Z})$ denote that $\mathbf{X}$ is separated from $\mathbf{Y}$ given $\mathbf{Z}$ in a graph $G$. Specifically, $sep(\mathbf{X}, \mathbf{Y}|\mathbf{Z})$ holds when every path in $G$ between $\mathbf{X}$ and $\mathbf{Y}$ is blocked by $\mathbf{Z}$. If $G$ is an undirected graph (UG), then a path in $G$ between $\mathbf{X}$ and $\mathbf{Y}$ is blocked by $\mathbf{Z}$ when there exists some $Z \in \mathbf{Z}$ in the path. We say that a node is a head-to-head node in a path if it has two parents in the path. If $G$ is a directed and acyclic graph (DAG), then a path in $G$ between $\mathbf{X}$ and $\mathbf{Y}$ is blocked by $\mathbf{Z}$ when there exists a node $Z$ in the path such that either (i) $Z$ is not a head-to-head node in the path and $Z \in \mathbf{Z}$, or (ii) $Z$ is a head-to-head node in the path and neither $Z$ nor any of its descendants in $G$ is in $\mathbf{Z}$. An independence model $p$ is faithful to an UG or DAG $G$ when $\mathbf{X} \perp\!\!\!\perp \mathbf{Y}|\mathbf{Z}$ iff $sep(\mathbf{X}, \mathbf{Y}|\mathbf{Z})$. Any independence model that is faithful to some UG or DAG is a graphoid. An UG (resp. DAG) $G$ is an undirected (resp. directed) independence map of an independence model $p$ when $\mathbf{X} \perp\!\!\!\perp \mathbf{Y}|\mathbf{Z}$ if $sep(\mathbf{X}, \mathbf{Y}|\mathbf{Z})$. Moreover, $G$ is a minimal undirected (resp. directed) independence map of $p$ when removing any edge from $G$ makes it cease to be an independence map of $p$. We abbreviate minimal undirected (resp. directed) independence map as MUI (resp. MDI) map. If $G$ is a MUI map of $p$, then two nodes $X$ and $Y$ are adjacent in $G$ iff $X \not\!\perp\!\!\!\perp Y|\mathbf{U} \setminus (XY)$. On the other hand, if $G$ is a MDI map of $p$, then the parents of a node $X$ in $G$, $Pa(X)$, are the smallest subset of the nodes preceding $X$ in a given total ordering of $\mathbf{U}$, $Pre(X)$, such that $X \perp\!\!\!\perp Pre(X) \setminus Pa(X) | Pa(X)$. We denote the children of a node $X$ by $Ch(X)$.

Let $Cl$ denote the set of cliques of an UG $G$. A Markov network (MN) is a pair $(G, \theta)$ where $G$ is an UG and $\theta$ are parameters specifying a non-negative function for the random variables in each $\mathbf{C} \in Cl$, $\phi(\mathbf{C})$. The MN represents the probability distribution $p = K \prod_{\mathbf{C} \in Cl} \phi(\mathbf{C})$ where $K$ is a normalizing constant. If a probability distribution $p$ can be represented by a MN with UG $G$, then $G$ is an undirected independence map of $p$. When $p$ is strictly positive, the opposite also holds.

A Bayesian network (BN) is a pair $(G, \theta)$ where $G$ is a DAG and $\theta$ are parameters specifying a multinomial probability distribution for each $X \in \mathbf{U}$ given its parents in $G$, $p(X|Pa(X))$. The BN represents the multinomial probability distribution $p = \prod_{X \in \mathbf{U}} p(X|Pa(X))$. A probability distribution $p$ can be represented by a BN with DAG $G$ iff $G$ is an directed independence map of $p$.

Given an UG (resp. DAG) $G$, we denote by $\mathcal{M}(G)$ all the multinomial probability distributions that can be represented by a MN (resp. BN) with UG (resp. DAG) $G$. Finally, recall that a polytree is a directed graph without undirected cycles. In other words, there exists at most one undirected path between any two nodes $X$ and $Y$. We denote that path by $X : Y$ if it exists. Recall also that a directed tree is a polytree where every node has at most one parent.

## 3 CWT GRAPHOIDS

Let $\mathbf{X}$, $\mathbf{Y}$, $\mathbf{Z}$ and $\mathbf{W}$ denote four mutually disjoint subsets of $\mathbf{U}$. We call CWT graphoid to any graphoid that satisfies composition $\mathbf{X} \perp\!\!\!\perp \mathbf{Y}|\mathbf{Z} \wedge \mathbf{X} \perp\!\!\!\perp \mathbf{W}|\mathbf{Z} \Rightarrow \mathbf{X} \perp\!\!\!\perp \mathbf{YW}|\mathbf{Z}$, and weak transitivity $\mathbf{X} \perp\!\!\!\perp \mathbf{Y}|\mathbf{Z} \wedge \mathbf{X} \perp\!\!\!\perp \mathbf{Y}|\mathbf{Z}V \Rightarrow \mathbf{X} \perp\!\!\!\perp V|\mathbf{Z} \vee V \perp\!\!\!\perp \mathbf{Y}|\mathbf{Z}$ with $V \in \mathbf{U} \setminus (\mathbf{XYZ})$. We now argue that there exist important families of probability distributions that are CWT graphoids and, thus, that WT graphoids are worth studying. For instance, any probability distribution that is Gaussian or faithful to some UG or DAG is a CWT graphoid [Pearl, 1988, Studený, 2005]. The theorem below proves that the marginals and conditionals of a probability distribution that is a CWT graphoid are CWT graphoids, although they may be neither Gaussian nor faithful to any UG or DAG. We give an example at the end of this section. We refer the reader to [Peña et al., 2006a, Peña et al., 2006b, Peña et al., 2007] for the proofs of the theorems in this section.

**Theorem 1** *Let $p$ be a probability distribution that is a CWT graphoid and let $\mathbf{W} \subseteq \mathbf{U}$. Then, $p(\mathbf{U} \setminus \mathbf{W})$ is a CWT graphoid. If $p(\mathbf{U} \setminus \mathbf{W}|\mathbf{W} = \mathbf{w})$ has the same independencies for all $\mathbf{w}$, then $p(\mathbf{U} \setminus \mathbf{W}|\mathbf{W} = \mathbf{w})$ for any $\mathbf{w}$ is a CWT graphoid.*

Hereinafter, we say that a probability distribution $p$ has context-specific independencies if there exists some $\mathbf{W} \subseteq \mathbf{U}$ such that $p(\mathbf{U} \setminus \mathbf{W}|\mathbf{W} = \mathbf{w})$ does not have the same independencies for all $\mathbf{w}$. We now show that it is not too restrictive to assume, as in the theorem above, that a probability distribution is a CWT graphoid that has no context-specific independencies, because there exist important families of probability distributions whose all or almost all the members satisfy such assumptions. For instance, a Gaussian probability distribution is a CWT graphoid [Studený, 2005], and has no context-specific independencies because its independencies are determined by its covariance matrix [Lauritzen, 1996]. The theorems below prove that this is also the case for almost all the probability distributions in $\mathcal{M}(G)$ where $G$ is an UG or DAG.

**Theorem 2** *Let $G$ be an UG. $\mathcal{M}(G)$ has non-zero Lebesgue measure wrt $\mathbb{R}^n$, where $n$ is the number of MN parameters corresponding to $G$. The probability distributions in $\mathcal{M}(G)$ that are not faithful to $G$ or have context-specific independencies have zero Lebesgue measure wrt $\mathbb{R}^n$.*

**Theorem 3** *Let $G$ be a DAG. $\mathcal{M}(G)$ has non-zero Lebesgue measure wrt $\mathbb{R}^n$, where $n$ is the number of BN parameters corresponding to $G$. The probability distributions in $\mathcal{M}(G)$ that are not faithful to $G$ or have context-specific independencies have zero Lebesgue measure wrt $\mathbb{R}^n$.*

The theorems above imply that, for any UG or DAG $G$, almost all the probability distributions in $\mathcal{M}(G)$ are CWT graphoids [Pearl, 1988], and that all the marginals and conditionals of almost all the probability dis-tributions in $\mathcal{M}(G)$ are CWT graphoids as well due to Theorem 1.

Finally, we give an example of a probability distribution that is a CWT graphoid although it is neither Gaussian nor faithful to any UG or DAG. Let $p$ be a multinomial probability distribution that is faithful to the DAG $\{X \to Y, Y \to Z, Z \to W, X \to V, V \to W, A \to B, C \to B\}$ and that has no context-specific independencies. Such a probability distribution exists due to Theorem 3 and, moreover, it is a CWT graphoid [Pearl, 1988]. Then, $p(X, Y, Z, V, A, B, C|W = w)$ for any $w$ is a CWT graphoid by Theorem 1, but it is neither Gaussian nor faithful to any UG or DAG, because it is multinomial and faithful to the graph $\{X - Y, Y - Z, Z - V, V - X, A \to B, C \to B\}$ [Chickering & Meek, 2002, Peña et al., 2006b].

## 4 READING DEPENDENCIES

In this section, we propose a sound and complete criterion for reading dependencies from a polytree-like MDI map of a CWT graphoid. If $G$ is a MDI map of a CWT graphoid $p$ then we know, by construction of $G$, that $X \not\!\perp\!\!\!\perp (Pre(X) \setminus Pa(X))Y|Pa(X) \setminus Y$ for all $X \in \mathbf{U}$ and $Y \in Pa(X)$. We call these dependencies the dependence base of $p$ for $G$. Further dependencies in $p$ can be derived from this dependence base via the CWT graphoid properties. For this purpose, we rephrase the CWT graphoid properties as follows. Let $\mathbf{X}$, $\mathbf{Y}$, $\mathbf{Z}$ and $\mathbf{W}$ denote four mutually disjoint subsets of $\mathbf{U}$. Symmetry $\mathbf{Y} \not\!\perp\!\!\!\perp \mathbf{X}|\mathbf{Z} \Rightarrow \mathbf{X} \not\!\perp\!\!\!\perp \mathbf{Y}|\mathbf{Z}$. Decomposition $\mathbf{X} \not\!\perp\!\!\!\perp \mathbf{Y}|\mathbf{Z} \Rightarrow \mathbf{X} \not\!\perp\!\!\!\perp \mathbf{YW}|\mathbf{Z}$. Weak union $\mathbf{X} \not\!\perp\!\!\!\perp \mathbf{Y}|\mathbf{ZW} \Rightarrow \mathbf{X} \not\!\perp\!\!\!\perp \mathbf{YW}|\mathbf{Z}$. Contraction $\mathbf{X} \not\!\perp\!\!\!\perp \mathbf{YW}|\mathbf{Z} \Rightarrow \mathbf{X} \not\!\perp\!\!\!\perp \mathbf{Y}|\mathbf{ZW} \vee \mathbf{X} \not\!\perp\!\!\!\perp \mathbf{W}|\mathbf{Z}$ is problematic for deriving new dependencies because it contains a disjunction in the right-hand side and, thus, it should be split into two properties: Contraction1 $\mathbf{X} \not\!\perp\!\!\!\perp \mathbf{YW}|\mathbf{Z} \wedge \mathbf{X} \perp\!\!\!\perp \mathbf{Y}|\mathbf{ZW} \Rightarrow \mathbf{X} \not\!\perp\!\!\!\perp \mathbf{W}|\mathbf{Z}$, and contraction2 $\mathbf{X} \not\!\perp\!\!\!\perp \mathbf{YW}|\mathbf{Z} \wedge \mathbf{X} \perp\!\!\!\perp \mathbf{W}|\mathbf{Z} \Rightarrow \mathbf{X} \not\!\perp\!\!\!\perp \mathbf{Y}|\mathbf{ZW}$. Likewise, intersection $\mathbf{X} \not\!\perp\!\!\!\perp \mathbf{YW}|\mathbf{Z} \Rightarrow \mathbf{X} \not\!\perp\!\!\!\perp \mathbf{Y}|\mathbf{ZW} \vee \mathbf{X} \not\!\perp\!\!\!\perp \mathbf{W}|\mathbf{ZY}$ gives rise to intersection1 $\mathbf{X} \not\!\perp\!\!\!\perp \mathbf{YW}|\mathbf{Z} \wedge \mathbf{X} \perp\!\!\!\perp \mathbf{Y}|\mathbf{ZW} \Rightarrow \mathbf{X} \not\!\perp\!\!\!\perp \mathbf{W}|\mathbf{ZY}$, and intersection2 $\mathbf{X} \not\!\perp\!\!\!\perp \mathbf{YW}|\mathbf{Z} \wedge \mathbf{X} \perp\!\!\!\perp \mathbf{W}|\mathbf{ZY} \Rightarrow \mathbf{X} \not\!\perp\!\!\!\perp \mathbf{Y}|\mathbf{ZW}$. Note that intersection1 and intersection2 are equivalent and, thus, we refer to them simply as intersection. Likewise, composition $\mathbf{X} \not\!\perp\!\!\!\perp \mathbf{YW}|\mathbf{Z} \Rightarrow \mathbf{X} \not\!\perp\!\!\!\perp \mathbf{Y}|\mathbf{Z} \vee \mathbf{X} \not\!\perp\!\!\!\perp \mathbf{W}|\mathbf{Z}$ gives rise to composition1 $\mathbf{X} \not\!\perp\!\!\!\perp \mathbf{YW}|\mathbf{Z} \wedge \mathbf{X} \perp\!\!\!\perp \mathbf{Y}|\mathbf{Z} \Rightarrow \mathbf{X} \not\!\perp\!\!\!\perp \mathbf{W}|\mathbf{Z}$, and composition2 $\mathbf{X} \not\!\perp\!\!\!\perp \mathbf{YW}|\mathbf{Z} \wedge \mathbf{X} \perp\!\!\!\perp \mathbf{W}|\mathbf{Z} \Rightarrow \mathbf{X} \not\!\perp\!\!\!\perp \mathbf{Y}|\mathbf{Z}$. Since composition1 and composition2 are equivalent, we refer to them simply as composition. Finally, weak transitivity $\mathbf{X} \not\!\perp\!\!\!\perp V|\mathbf{Z} \wedge V \not\!\perp\!\!\!\perp \mathbf{Y}|\mathbf{Z} \Rightarrow \mathbf{X} \not\!\perp\!\!\!\perp \mathbf{Y}|\mathbf{Z} \vee \mathbf{X} \not\!\perp\!\!\!\perp \mathbf{Y}|\mathbf{Z}V$ with $V \in \mathbf{U} \setminus (\mathbf{XYZ})$ gives rise to weak transitivity1 $\mathbf{X} \not\!\perp\!\!\!\perp V|\mathbf{Z} \wedge V \not\!\perp\!\!\!\perp \mathbf{Y}|\mathbf{Z} \wedge \mathbf{X} \perp\!\!\!\perp \mathbf{Y}|\mathbf{Z} \Rightarrow \mathbf{X} \not\!\perp\!\!\!\perp \mathbf{Y}|\mathbf{Z}V$, and weak transitivity2 $\mathbf{X} \not\!\perp\!\!\!\perp V|\mathbf{Z} \wedge V \not\!\perp\!\!\!\perp \mathbf{Y}|\mathbf{Z} \wedge \mathbf{X} \perp\!\!\!\perp \mathbf{Y}|\mathbf{Z}V \Rightarrow \mathbf{X} \not\!\perp\!\!\!\perp \mathbf{Y}|\mathbf{Z}$. The independence in the left-hand side of any of the properties above holds if the corresponding $sep$ statement holds in $G$. This is the best solution we can hope for because $sep$ is sound and complete. By definition, $sep$ is sound in the sense that it only identifies independencies in $p$. Furthermore, $sep$ is complete in the sense that it identifies all the independencies in $p$ that can be identified by studying $G$ alone because (i) there exist multinomial and Gaussian probability distributions that are faithful to $G$ [Meek, 1995], and (ii) such probability distributions are CWT graphoids [Pearl, 1988], $G$ is a MDI map for them, and they only have the independencies that $sep$ identifies from $G$. Moreover, this solution does not require more information about $p$ than what it is available, because $G$ can be constructed from the dependence base of $p$ for $G$. We call the CWT graphoid closure of the dependence base of $p$ for $G$ to the set of dependencies that are in the dependence base of $p$ for $G$ plus those that can be derived from it by applying the CWT graphoid properties.

We now introduce our criterion for reading dependencies from a polytree-like MDI map $G$ of a CWT graphoid $p$.

**Definition 1** *Let $\mathbf{X}$, $\mathbf{Y}$ and $\mathbf{Z}$ denote three mutually disjoint subsets of $\mathbf{U}$. We say that $dep(\mathbf{X}, \mathbf{Y}|\mathbf{Z})$ holds if there exist some $A \in \mathbf{X}$ and $B \in \mathbf{Y}$ such that (i) $sep(A, B|\mathbf{Z})$ does not hold, and (ii) for every head-to-head node $C$ in $A : B$, either $\mathbf{Z}$ contains $C$ or $\mathbf{Z}$ contains exactly one descendant of $C$ that is not a descendant of another descendant of $C$ that is in $\mathbf{Z}$.*

We now prove that $dep$ is sound, i.e. it only identifies dependencies in $p$. Actually, it only identifies dependencies in the CWT graphoid closure of the dependence base of $p$ for $G$.

**Theorem 4** *Let $G$ be a MDI map of a CWT graphoid $p$. If $G$ is a polytree then, if $dep(\mathbf{X}, \mathbf{Y}|\mathbf{Z})$ then $\mathbf{X} \not\!\perp\!\!\!\perp \mathbf{Y}|\mathbf{Z}$ is in the CWT graphoid closure of the dependence base of $p$ for $G$.*

**Proof:** Let us assume that every head-to-head node in $A : B$ is in $\mathbf{Z}$. We prove that $dep(A, B|\mathbf{Z})$ implies $A \not\!\perp\!\!\!\perp B|\mathbf{Z}$. We prove it by induction over the length of $A : B$.

We first prove the result for length one, i.e. $A : B$ is $A \to B$ or $A \leftarrow B$. Assume without loss of generality that $A : B$ is $A \to B$. Let $\mathbf{Z}^A$ denote the nodes in $\mathbf{Z}$ that are in $Pa(A)$ or connected to $A$ by an undirected path that passes through $Pa(A)$. Let $\mathbf{Z}_A$ denote the nodes in $\mathbf{Z}$ that are in $Ch(A) \setminus B$ or connected to $A$ by an undirected path that passes through $Ch(A) \setminus B$. Let $\mathbf{Z}^B$ denote the nodes in $\mathbf{Z}$ that are in $Pa(B) \setminus A$ or connected to $B$ by an undirected path that passes through $Pa(B) \setminus A$. Let $\mathbf{Z}_B$ denote the nodes in $\mathbf{Z}$ that are in $Ch(B)$ or connected to $B$ by an undirected path that passes through $Ch(B)$. Note that the nodes in $\mathbf{Z} \setminus (\mathbf{Z}^A \mathbf{Z}_A \mathbf{Z}^B \mathbf{Z}_B)$ are not connected to $A$ or $B$ by any undirected path. Then,

(1) $A(Pre(B) \setminus Pa(B)) \not\perp B | Pa(B) \setminus A$ from the dependence base of $p$ for $G$

(2) $Pre(B) \setminus Pa(B) \perp B | Pa(B)$ by $sep(Pre(B) \setminus Pa(B), B | Pa(B))$

(3) $A \not\perp B | Pa(B) \setminus A$ by contraction1 on (1) and (2)

(4) $A \not\perp B(Pa(B) \setminus A) | \emptyset$ by weak union on (3)

(5) $A \perp Pa(B) \setminus A | \emptyset$ by $sep(A, Pa(B) \setminus A | \emptyset)$

(6) $A \not\perp B | \emptyset$ by composition on (4) and (5)

(7) $A \mathbf{Z}^A \mathbf{Z}_A \not\perp B | \emptyset$ by decomposition on (6)

(8) $\mathbf{Z}^A \mathbf{Z}_A \perp B | A$ by $sep(\mathbf{Z}^A \mathbf{Z}_A, B | A)$

(9) $A \not\perp B | \mathbf{Z}^A \mathbf{Z}_A$ by intersection on (7) and (8)

(10) $A \not\perp B \mathbf{Z}^B | \mathbf{Z}^A \mathbf{Z}_A$ by decomposition on (9)

(11) $A \perp \mathbf{Z}^B | \mathbf{Z}^A \mathbf{Z}_A$ by $sep(A, \mathbf{Z}^B | \mathbf{Z}^A \mathbf{Z}_A)$

(12) $A \not\perp B | \mathbf{Z}^A \mathbf{Z}_A \mathbf{Z}^B$ by contraction2 on (10) and (11)

(13) $A \not\perp B \mathbf{Z}_B | \mathbf{Z}^A \mathbf{Z}_A \mathbf{Z}^B$ by decomposition on (12)

(14) $A \perp \mathbf{Z}_B | \mathbf{Z}^A \mathbf{Z}_A \mathbf{Z}^B B$ by $sep(A, \mathbf{Z}_B | \mathbf{Z}^A \mathbf{Z}_A \mathbf{Z}^B B)$

(15) $A \not\perp B | \mathbf{Z}^A \mathbf{Z}_A \mathbf{Z}^B \mathbf{Z}_B$ by intersection on (13) and (14)

(16) $A \not\perp B \mathbf{Z} \setminus (\mathbf{Z}^A \mathbf{Z}_A \mathbf{Z}^B \mathbf{Z}_B) | \mathbf{Z}^A \mathbf{Z}_A \mathbf{Z}^B \mathbf{Z}_B$ by decomposition on (15)

(17) $A \perp \mathbf{Z} \setminus (\mathbf{Z}^A \mathbf{Z}_A \mathbf{Z}^B \mathbf{Z}_B) | \mathbf{Z}^A \mathbf{Z}_A \mathbf{Z}^B \mathbf{Z}_B$ by $sep(A, \mathbf{Z} \setminus (\mathbf{Z}^A \mathbf{Z}_A \mathbf{Z}^B \mathbf{Z}_B) | \mathbf{Z}^A \mathbf{Z}_A \mathbf{Z}^B \mathbf{Z}_B)$

(18) $A \not\perp B | \mathbf{Z}$ by contraction2 on (16) and (17).

Assume as induction hypothesis that the result holds when the length of $A : B$ is smaller than $l$. We now prove the result for length $l$. Let $C$ be any node in $A : B$ except $A$ and $B$. If $C$ is not a head-to-head node in $A : B$, then

(19) $A \not\perp C | \mathbf{Z}$ by the induction hypothesis

(20) $B \not\perp C | \mathbf{Z}$ by the induction hypothesis

(21) $A \perp B | \mathbf{Z} C$ by $sep(A, B | \mathbf{Z} C)$

(22) $A \not\perp B | \mathbf{Z}$ by weak transitivity2 on (19-21).

On the other hand, if $C$ is a head-to-head node in $A : B$, then $C$ must be in $\mathbf{Z}$ due to the assumption made at the beginning of this proof. Let $\mathbf{Z}_C$ denote the nodes in $\mathbf{Z}$ that are in $Ch(C)$ or connected to $C$ by an undirected path that passes through $Ch(C)$. Then,

(23) $A \not\perp C | \mathbf{Z} \setminus (\mathbf{Z}_C C)$ by the induction hypothesis

(24) $B \not\perp C | \mathbf{Z} \setminus (\mathbf{Z}_C C)$ by the induction hypothesis

(25) $A \perp B | \mathbf{Z} \setminus (\mathbf{Z}_C C)$ by $sep(A, B | \mathbf{Z} \setminus (\mathbf{Z}_C C))$

(26) $A \not\perp B | \mathbf{Z} \setminus \mathbf{Z}_C$ by weak transitivity1 on (23-25).

(27) $A \not\perp B \mathbf{Z}_C | \mathbf{Z} \setminus \mathbf{Z}_C$ by decomposition on (26)

(28) $A \perp \mathbf{Z}_C | \mathbf{Z} \setminus \mathbf{Z}_C$ by $sep(A, \mathbf{Z}_C | \mathbf{Z} \setminus \mathbf{Z}_C)$

(29) $A \not\perp B | \mathbf{Z}$ by contraction2 on (27) and (28).

Let us now assume that for every head-to-head node $C$ in $A : B$, either $\mathbf{Z}$ contains $C$ or $\mathbf{Z}$ contains exactly one descendant of $C$ that is not a descendant of another descendant of $C$ that is in $\mathbf{Z}$. We prove that $dep(A, B | \mathbf{Z})$ implies $A \not\perp B | \mathbf{Z}$. We prove it by induction over the number of head-to-head nodes in $A : B$ that are not in $\mathbf{Z}$. As proven above, the result holds when this number is zero. Assume as induction hypothesis that the result holds when this number is smaller than $l$. We now prove the result for $l$. Let $C$ be any head-to-head node in $A : B$ that is not in $\mathbf{Z}$. Then, $\mathbf{Z}$ must contain exactly one descendant of $C$, say $D$, that is not a descendant of another descendant of $C$ that is in $\mathbf{Z}$. Let $\mathbf{Z}_D$ denote the nodes in $\mathbf{Z}$ that are in $Ch(D)$ or connected to $D$ by an undirected path that passes through $Ch(D)$. Then,

(30) $A \not\perp D | \mathbf{Z} \setminus (\mathbf{Z}_D D)$ by the induction hypothesis

(31) $B \not\perp D | \mathbf{Z} \setminus (\mathbf{Z}_D D)$ by the induction hypothesis

(32) $A \perp B | \mathbf{Z} \setminus (\mathbf{Z}_D D)$ by $sep(A, B | \mathbf{Z} \setminus (\mathbf{Z}_D D))$

(33) $A \not\perp B | \mathbf{Z} \setminus \mathbf{Z}_D$ by weak transitivity1 on (30-32)

(34) $A \not\perp B \mathbf{Z}_D | \mathbf{Z} \setminus \mathbf{Z}_D$ by decomposition on (33)

(35) $A \perp \mathbf{Z}_D | \mathbf{Z} \setminus \mathbf{Z}_D$ by $sep(A, \mathbf{Z}_D | \mathbf{Z} \setminus \mathbf{Z}_D)$

(36) $A \not\perp B | \mathbf{Z}$ by contraction2 on (34) and (35).

Therefore, we have proven that $dep(A, B | \mathbf{Z})$ implies $A \not\perp B | \mathbf{Z}$ which, in turn, implies $\mathbf{X} \not\perp \mathbf{Y} | \mathbf{Z}$ by decomposition. Since $dep(\mathbf{X}, \mathbf{Y} | \mathbf{Z})$ implies $dep(A, B | \mathbf{Z})$, $dep(\mathbf{X}, \mathbf{Y} | \mathbf{Z})$ implies $\mathbf{X} \not\perp \mathbf{Y} | \mathbf{Z}$. Moreover, this last dependence must be in the CWT graphoid closure of the dependence base of $p$ for $G$, because we have only used in the proof the dependence base of $p$ for $G$ and the CWT graphoid properties. $\square$

We now prove that $dep$ is complete for reading dependencies from a polytree-like MDI map $G$ of a CWT graphoid $p$, in the sense that it identifies all the dependencies in $p$ that follow from the information about $p$ that is available, namely the dependence base of $p$ for $G$ and the fact that $p$ is a CWT graphoid.

**Theorem 5** *Let $G$ be a MDI map of a CWT graphoid $p$. If $G$ is a polytree, then if $\mathbf{X} \not\perp \mathbf{Y} | \mathbf{Z}$ is in the CWT graphoid closure of the dependence base of $p$ for $G$ then $dep(\mathbf{X}, \mathbf{Y} | \mathbf{Z})$.*

**Proof:** It suffices to prove (i) that all the dependencies in the dependence base of $p$ for $G$ are identified by $dep$, and (ii) that $dep$ satisfies the CWT graphoid properties. Since the first point is trivial, we only prove the second point. Let $\mathbf{X}$, $\mathbf{Y}$, $\mathbf{Z}$ and $\mathbf{W}$ denote four mutually disjoint subsets of $\mathbf{U}$.

- Symmetry $dep(\mathbf{Y}, \mathbf{X}|\mathbf{Z}) \Rightarrow dep(\mathbf{X}, \mathbf{Y}|\mathbf{Z})$. The path $A : B$ in the left-hand side also satisfies the right-hand side.

- Decomposition $dep(\mathbf{X}, \mathbf{Y}|\mathbf{Z}) \Rightarrow dep(\mathbf{X}, \mathbf{YW}|\mathbf{Z})$. The path $A : B$ in the left-hand side also satisfies the right-hand side.

- Weak union $dep(\mathbf{X}, \mathbf{Y}|\mathbf{ZW}) \Rightarrow dep(\mathbf{X}, \mathbf{YW}|\mathbf{Z})$. The path $A : B$ in the left-hand side also satisfies the right-hand side unless there exists some head-to-head node $C$ in $A : B$ such that neither $C$ nor any of its descendants is in $\mathbf{Z}$ (if several such nodes exist, let $C$ be the closest to $A$). However, $C$ or some descendant $D$ of $C$ must be in $\mathbf{W}$ for $dep(\mathbf{X}, \mathbf{Y}|\mathbf{ZW})$ to hold. Then, $A : C$ or $A : D$ satisfies the right-hand side.

- Contraction1 $dep(\mathbf{X}, \mathbf{YW}|\mathbf{Z}) \wedge sep(\mathbf{X}, \mathbf{Y}|\mathbf{ZW}) \Rightarrow dep(\mathbf{X}, \mathbf{W}|\mathbf{Z})$. Let $C$ denote the closest node to $A$ that is in both $\mathbf{W}$ and the path $A : B$ in the left-hand side. Such a node must exist for $sep(\mathbf{X}, \mathbf{Y}|\mathbf{ZW})$ to hold. For the same reason, no node in $A : C$ can be in $\mathbf{Y}$. Then, $A : C$ satisfies the right-hand side.

- Contraction2 $dep(\mathbf{X}, \mathbf{YW}|\mathbf{Z}) \wedge sep(\mathbf{X}, \mathbf{W}|\mathbf{Z}) \Rightarrow dep(\mathbf{X}, \mathbf{Y}|\mathbf{ZW})$. Let $C$ denote the closest node to $A$ that is in both $\mathbf{Y}$ and the path $A : B$ in the left-hand side. Such a node must exist for $sep(\mathbf{X}, \mathbf{W}|\mathbf{Z})$ to hold. For the same reason, no node in $A : C$ can be in $\mathbf{W}$. Then, $A : C$ satisfies the right-hand side.

- Intersection $dep(\mathbf{X}, \mathbf{YW}|\mathbf{Z}) \wedge sep(\mathbf{X}, \mathbf{Y}|\mathbf{ZW}) \Rightarrow dep(\mathbf{X}, \mathbf{W}|\mathbf{ZY})$. Consider the same reasoning as in contraction1.

- Composition $dep(\mathbf{X}, \mathbf{YW}|\mathbf{Z}) \wedge sep(\mathbf{X}, \mathbf{W}|\mathbf{Z}) \Rightarrow dep(\mathbf{X}, \mathbf{Y}|\mathbf{Z})$. Consider the same reasoning as in contraction2.

- Weak transitivity1 $dep(\mathbf{X}, V|\mathbf{Z}) \wedge dep(V, \mathbf{Y}|\mathbf{Z}) \wedge sep(\mathbf{X}, \mathbf{Y}|\mathbf{Z}) \Rightarrow dep(\mathbf{X}, \mathbf{Y}|\mathbf{Z}V)$ with $V \in \mathbf{U} \setminus (\mathbf{XYZ})$. Let $A : V$ and $V : B$ denote the paths in the first and second, respectively, $dep$ statements in the left-hand side. For $sep(\mathbf{X}, \mathbf{Y}|\mathbf{Z})$ to hold, $V$ must be a head-to-head node in $A : B$ or the descendant of a head-to-head node in $A : B$ that neither is in $\mathbf{Z}$ nor has any descendant in $\mathbf{Z}$. Then, $A : B$ satisfies the right-hand side.

- Weak transitivity2 $dep(\mathbf{X}, V|\mathbf{Z}) \wedge dep(V, \mathbf{Y}|\mathbf{Z}) \wedge sep(\mathbf{X}, \mathbf{Y}|\mathbf{Z}V) \Rightarrow dep(\mathbf{X}, \mathbf{Y}|\mathbf{Z})$ with $V \in \mathbf{U} \setminus (\mathbf{XYZ})$. Let $A : V$ and $V : B$ denote the paths in the first and second, respectively, $dep$ statements in the left-hand side. Note that $V$ must be a non-head-to-head node in $A : B$ for $sep(\mathbf{X}, \mathbf{Y}|\mathbf{Z}V)$ to hold. Then, $A : B$ satisfies the right-hand side.

$\square$

Finally, the soundness of $dep$ allows us to prove the following theorem.

**Theorem 6** *Let $G$ be a MDI map of a CWT graphoid $p$. If $G$ is a directed tree, then $p$ is faithful to it.*

**Proof:** Any independence in $p$ for which the corresponding separation statement does not hold in $G$ contradicts Theorem 4. $\square$

The theorem above has been proven by [Becker et al., 2000] for the case where $G$ is a tree-like MUI map and $p$ is either a Gaussian probability distribution or a multinomial probability distribution over binary random variables. In [Peña et al., 2006a], we have proven the theorem above for the case where $G$ is a tree-like MUI map and $p$ is a graphoid that satisfies weak transitivity.

## 5 DISCUSSION

We have presented a sound and complete graphical criterion for reading dependencies from a polytree-like MDI map $G$ of a CWT graphoid. We have shown that there exist important families of probability distributions, among them Gaussian probability distributions, that are CWT graphoids. We think that this work complements previous works addressing the same question, e.g. [Bouckaert, 1995] which proposes sound and complete graphical criteria for reading dependencies from a MUI or MDI map of a graphoid, and [Peña et al., 2006a, Peña et al., 2006b, Peña et al., 2007] which propose a sound and complete graphical criterion for reading dependencies from the MUI map of graphoid that satisfies weak transitivity. See also [Peña et al., 2007] for a real-world application of the graphical criterion developed in that paper to read biologically meaningful gene dependencies.

Our end-goal is to apply the results in this paper to our project on atherosclerosis gene expression data analysis in order to learn dependencies between genes. We believe that it is not unrealistic to assume that the probability distribution underlying our data is a CWT graphoid. We base this belief on the following argument. The cell is the functional unit of all the organisms and includes all the information necessary to regulate its function. This information is encoded in the DNA of the cell, which is divided into a set of genes, each coding for one or more proteins. Proteins are required for practically all the functions in the cell. The amount of protein produced depends on the expression level of the coding gene which, in turn, depends on the amount of proteins produced by other genes. Therefore, a dynamic BN or Gaussian network (GN) seems to be a relatively accurate model of the cell: The nodes represent the genes and proteins, and the edges and parameters represent the causal relations between the gene expression levels and the protein amounts. As a matter of fact, dynamic BNs and GNs have become very popular models of gene networks for the last few years [Bernard & Hartemink, 2005, Friedman et al., 1998, Husmeier, 2003, Kim et al., 2003, Murphy & Mian, 1999, Ong et al., 2002, Perrin et al., 2003, Zou & Conzen, 2005]. It is important that the BN or GN is dynamic because a gene can regulate some of its regulators and even itself with some time delay. Since the technology for measuring the state of the protein nodes is not widely available yet, the data in most projects on gene expression data analysis is a finite sample of the probability distribution represented by the dynamic BN or GN after marginalizing the protein nodes out. The probability distribution with no node marginalized out is almost surely faithful to the dynamic BN or GN [Meek, 1995] and, thus, it satisfies composition and weak transitivity (see Section 3) and, thus, so does the probability distribution after marginalizing the protein nodes out (see Theorem 1).

Other graphical models that have received increasing attention from the bioinformatics community as a means to gain insight into gene networks are the so-called graphical Gaussian models [Castelo & Roverato, 2006, Dobra et al., 2004, Kishino & Waddell, 2000, Li & Gui, 2006, Schäfer & Strimmer, 2005a, Schäfer & Strimmer, 2005b, Toh & Horimoto, 2002, Waddell & Kishino, 2000, Wang et al., 2003, Wu et al., 2003]. A graphical Gaussian model is nothing else but the MUI map of the probability distribution underlying the gene network under the assumption that this is Gaussian and, thus, a CWT graphoid. This again supports our belief that the theory developed in this paper may be relevant within bioinformatics.

## Acknowledgements

This work is funded by the Swedish Research Council (ref. VR-621-2005-4202). We thank the three anonymous referees for their comments.